\documentclass[10pt,twocolumn,letterpaper]{article}

\usepackage{cvpr}              % To produce the CAMERA-READY version
\definecolor{cvprblue}{rgb}{0.21,0.49,0.74}
\usepackage[pagebackref,breaklinks,colorlinks,allcolors=cvprblue]{hyperref}

\def\paperID{43} % *** Enter the Paper ID here
\def\confName{CVPR}
\def\confYear{2025}

\title{Training-Free Sketch-Guided Diffusion with Latent Optimization}

\author{Sandra Zhang Ding \ \ \ \ Jiafeng Mao \ \ \ \  Kiyoharu Aizawa\\
The University of Tokyo\\
{\tt\small \{sandra, mao, aizawa\}@hal.t.u-tokyo.ac.jp}
}
\begin{document}
\maketitle
%\input{sec/0_abstract}    
%\input{sec/1_intro}
%\input{sec/2_formatting}
%\input{sec/3_finalcopy}

%%%%%%%%% ABSTRACT
\begin{abstract}
Based on recent advanced diffusion models, Text-to-image (T2I) generation models have demonstrated their capabilities to generate diverse and high-quality images.
However, leveraging their potential for real-world content creation, particularly in providing users with precise control over the image generation result, poses a significant challenge.
In this paper, we propose an innovative training-free pipeline that extends existing text-to-image generation models to incorporate a sketch as an additional condition.
To generate new images with a layout and structure closely resembling the input sketch, we find that these core features of a sketch can be tracked with the cross-attention maps of diffusion models.
We introduce latent optimization, a method that refines the noisy latent at each intermediate step of the generation process using cross-attention maps to ensure that the generated images adhere closely to the desired structure outlined in the reference sketch.
Through latent optimization, our method enhances the accuracy of image generation, offering users greater control and customization options in content creation.
\end{abstract}

%%%%%%%%% BODY TEXT
\vspace{-10pt}
\section{Introduction}
\label{sec:introduction}

\begin{figure}
    \centering
    \includegraphics[width=0.95\linewidth]{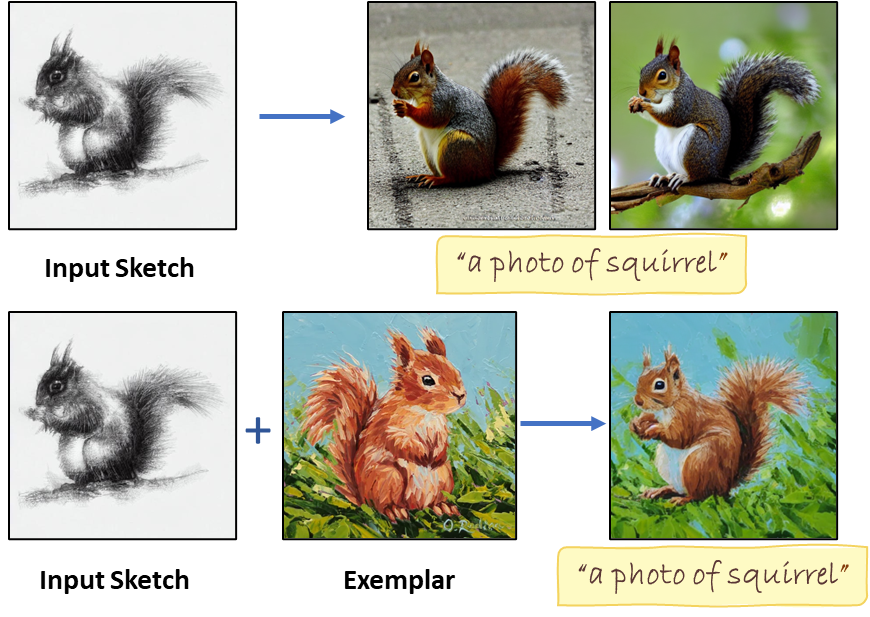}
    \caption{
    Given a sketch and a text prompt, our pipeline synthesizes an image that adheres to the sketch structure and the text description. If the user wants to use an additional exemplar image as another input, we can also perform image variation while maintaining fidelity to the sketch.}
    \label{fig:teaser}
    \vspace{-10pt}
\end{figure}
% intro
Image generation constitutes a crucial domain within computer vision research.
Diffusion models~\cite{ho2020denoising, nichol2021improved, nichol2021glide, saharia2022photorealistic} have emerged as promising tools for generating high-fidelity images through the iterative denoising of pure Gaussian noise inputs.
Previous research~\cite{rombach2022high, ramesh2022hierarchical, dhariwal2021diffusion} in this realm has concentrated on employing prompts to steer image generation.
However, designing optimal prompts for desired content poses a notable challenge.
Consequently, various methods~\cite{chefer2023attend, lugmayr2022repaint, avrahami2022blended, ruiz2023dreambooth, horita2022structure} have been proposed to facilitate fine-grained control over the image generation process.

% what we did 
Sketch is a straightforward yet potent medium for users to convey their ideas visually. 
With just a few strokes, these drawings can translate abstract concepts into tangible visual narratives, offering an intuitive medium to depict various ideas and concepts.
However, converting sketches to realistic images is challenging due to the domain gap between realistic images and sketch images.

%sketch-to-image task
To achieve the training-free sketch-guided image generation task, we focus on the strong visual comprehension capabilities of pre-trained diffusion models.
Without additional training, the diffusion models can recognize and extract high-level features from images across different domains when provided with prompts containing the relevant domain information.
Specifically, we find that the layout and structural features preserved within the noise latents of diffusion models can be monitored by cross-attention maps, which enables these maps to be used as guidance for the generation of realistic images.
In our method, we perform a DDIM Inversion~\cite{song2020denoising} on the reference sketch provided by the user, preserving the model's internal responses, i.e., cross-attention maps containing the sketch features, at each denoising step of the reconstruction process.
Then, we perform image generation using a randomly initialized noise with the guidance of the textual prompt given by the user. 
To generate images that match the input sketch, we propose a technique called latent optimization and apply it during the generation process.
At each intermediate denoising step of realistic images, we treat the noisy latents as variables and optimize them by aligning the internal response with the saved attention maps obtained during DDIM Inversion of the sketch input.
By optimizing the noise latent in this manner, we ensure that the generated images closely adhere to the desired structure in the reference sketch.

We experimentally evaluate the effectiveness of our method using two distinct datasets: (1) the Sketchy database~\cite{sketchy2016} and (2) the ImageNet-Sketch dataset~\cite{wang2019learning}.
The Sketchy database consists exclusively of highly abstract scribble sketches, whereas the ImageNet-Sketch dataset includes sketches with varying levels of abstraction, ranging from highly detailed edge maps to more simplified line-art representations.
The fact that no training is required allows our approach to be applied to reference sketches from different sketch domains. 
Whether using an extremely abstract sketch or one with a large amount of detail as a reference image, our training-free method effectively generates the image as intended.
This result confirms that the diffusion model has a robust graphical understanding and can recognize and extract key object features of sketches from different domains.
We further extend our approach to real image editing, where the model receives both the real image and the target sketch, and outputs a variation of the real image according to the guidance of the sketch.
Our experiments show that our method can effectively align sketch guidance while preserving the content of the original image.

% contributions
Our contributions can be summarized as follows:
%\vspace{-0.5\baselineskip}
\begin{itemize}
\setlength{\itemsep}{0pt}
\setlength{\parskip}{0pt}
\item We find the strong visual discrimination capabilities reside within diffusion models, where the layout and structure features of sketch inputs can be preserved in the cross-attention maps of the diffusion model.
%\item  We propose latent optimization, a technique used to refine intermediate noisy latent to align the features between the reference and generated images for controllable image generation. 
\item  We propose latent optimization to refine intermediate noisy latent to align the features between the reference and generated images for controllable image generation. 
\item The proposed method is effective for sketch-to-image generation without the need for additional training or fine-tuning. Moreover, it proves successful in editing real images based on sketch guidance.
\end{itemize}

\section{Related Work}
\label{sec:related-work}

\subsection{Text-to-Image Synthesis and Editing}
\label{subsec:image-editing}
Generative diffusion models are capable of producing image samples from Gaussian noise through an iterative noise removal process. 
The recent emergence of diffusion models trained on large-scale image-text datasets has further propelled advancements in image generation~\cite{nichol2021glide, kim2022diffusionclip, ho2022classifier, saharia2022photorealistic}. 
These models, leveraging the power of the text encoder~\cite{radford2021learning, raffel2020exploring}, have facilitated the integration of text as a versatile handler for image generation. 
Leveraging the power of text-to-image models, several approaches have been proposed to manipulate images globally or locally using text.
In~\cite{chen2024training, hertz2022prompt, brooks2023instructpix2pix, mokady2023null, parmar2023zero, balaji2022ediffi, mao2023training, mao2023guided, mao2024lottery},  by manipulating cross-attention maps, it becomes possible to achieve flexible image generation and editing, such as altering local objects or modifying global image styles. 
However, these approaches face challenges in modifying the fine-grained object attributes of real images due to the abstract nature of the text. 
To address this limitation, our approach builds upon Stable Diffusion~\cite{rombach2022high} and incorporates sketches as an intuitive and precise control signal for image manipulation.

%\vspace{-2pt}
\subsection{Sketch-based Image Synthesis}
%\vspace{-3pt}
\label{subsec:sketch-based}
% add GAN (Sketch Your Own GAN if there is space)
Several methods~\cite{voynov2023sketch, mou2023t2i, zhang2023adding, peng2023sketch, cheng2023adaptively, koley2024handle} have been proposed to perform the sketch-to-image synthesis task by training additional networks.
These methods are capable of transforming abstract inputs, such as edge maps, into realistic images.
However, they require extra data and training, which limits their scalability and accessibility.
Moreover, their performance often depends heavily on the type and quality of the sketch.
When users provide only rough and crude sketches, these methods tend to generate content of lower quality, as they rely heavily on the detail and accuracy of the input sketch.

In contrast, training-free methods~\cite{tumanyan2023plug, hertz2022prompt} perform tasks by leveraging guidance from a reference image. 
Yet, their performance declines when the guidance is a sketch, due to the domain gap between sketches and real images.
Unlike existing approaches, our method not only preserves positional alignment but also reconstructs the core features of the input sketch in the generated image—achieving high fidelity without requiring additional training.

\section{Proposed Method}
\label{sec:approach}

In this section, we present our training-free method for the sketch-to-image generation task. 
Our approach is built entirely on a single pre-trained text-to-image model, \ie, Stable Diffusion (SD)~\cite{rombach2022high}, a widely used model that performs denoising in the latent space rather than the image space.
We first outlines the preliminary techniques of diffusion models used in our method in Sec.~\ref{sec:preliminary}.
%Then, in Section \ref{sec:noise_latent}, we discuss our novel observation and the motivation for the proposed method. Despite a gap between realistic and sketchy images in inverted latents obtained from DDIM Inversion\cite{song2020denoising}, cross-attention maps effectively track layout and structural features during the reconstruction process across different domains.
Next, in Section~\ref{sec:noise_latent}, we present a key observation and the motivation behind our approach: although DDIM Inversion~\cite{song2020denoising} reveals a clear domain gap between real and sketch images in the latent space, cross-attention maps still successfully capture and track structural and layout features during reconstruction—regardless of domain.
Building on this, we propose a latent optimization strategy that incorporates these attention maps into the generation process to guide the model in producing outputs that adhere to the structure of the reference sketch, as detailed in Section~\ref{sec:pipeline}.

\begin{figure*}[t]
    \centering
    \includegraphics[width=1.0\linewidth]{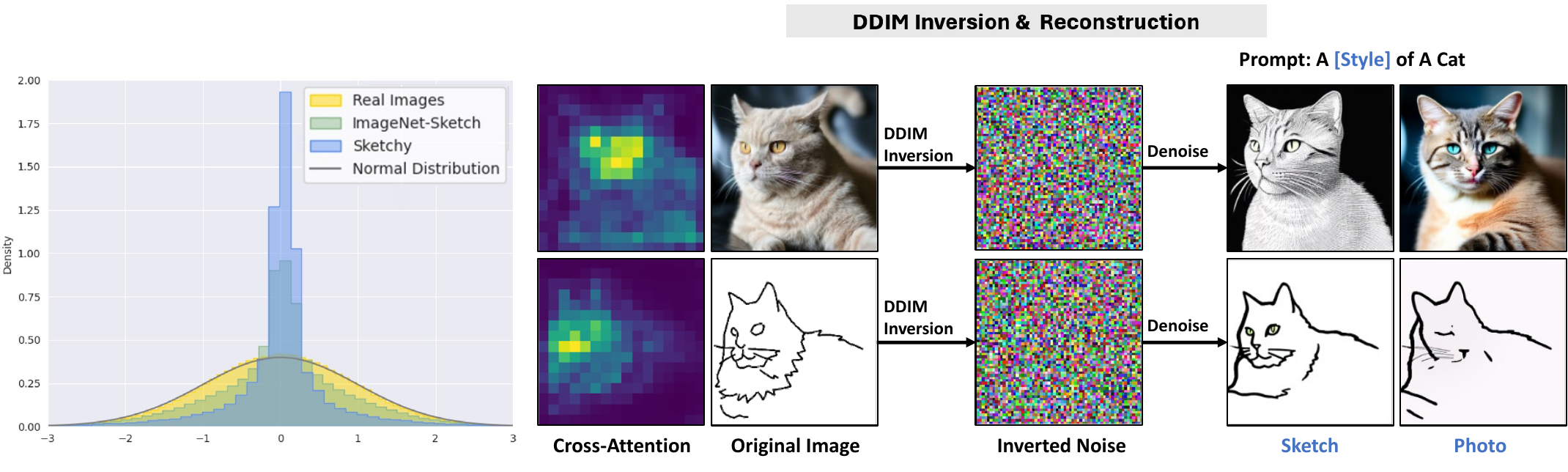}
    %\vspace{-10pt}
    \caption{
    Left: Distributions of inverted sketches (ImageNet-Sketch and Sketchy) show noticeable bias compared to the standard normal distribution and those of inverted real images.
    Right: Comparison of using different initial noises with different prompts. Using an inverted sketch image generates a sketch-like image even when using the style keyword \texttt{"photo"}. Note that the locations of the cats in the generated image align with the highlighted areas in the attention maps.}%结论
    \label{fig:initial-generation-tendency}
    \vspace{-15pt}
\end{figure*}

\subsection{Preliminary}
\label{sec:preliminary}
\subsubsection{DDIM Inversion}
\label{subsubsec:ddim-inversion}
As widely recognized, the diffusion models act as time-dependent decoders, denoted by $\epsilon_\theta(z_t, t)$.
These decoders iteratively refine the latent representation $z_t$ for $ t = 1, 2, ..., T $, beginning with an initial Gaussian noise latent $z_T$.
Denoising diffusion implicit models (DDIMs)~\cite{song2020denoising} enable deterministic and approximately invertible image generation. 
By reversing the DDIM sampling process, one can perform DDIM inversion to obtain the initial noise latent $z_T$ from a real image, which can then be denoised to reconstruct the original image.
This inversion technique has been widely adopted in image editing and image-to-image translation tasks~\cite{tumanyan2023plug, parmar2023zero, mokady2023null, hertz2022prompt}. 
By generating from the noise latent obtained via DDIM inversion, these methods preserve key features of a reference image while enabling additional guidance for editing purposes.
%\vspace{-5pt}

\subsubsection{Cross-Attention Mechanism}
\label{sec:cross-attention}

In recent diffusion models~\cite{rombach2022high, peebles2023scalable} as Stable Diffusion, the U-Net backbone~\cite{ronneberger2015u} is augmented with a cross-attention mechanism~\cite{vaswani2017attention} to incorporate additional conditions, such as text, into the image generation process.
The model uses a CLIP encoder~\cite{radford2021learning} to convert the input text into a text embedding $c$. During the denoising process, cross-attention maps are computed to align and integrate the text semantics with the intermediate spatial features of the image, as defined by:
\begin{equation}
    M = \text{Softmax}(\frac{QK^T}{\sqrt{d_k}}),
    \label{eq:attention}
\end{equation} 
where the query $ Q=W_Q\cdot{\tau(z_t)} $ is derived from the intermediate spatial features $\tau(z_t)$ of the denoising network, and the key $K=W_K\cdot{c} $ comes from the text embedding $c$, both applied by learned weight matrices $W^{}_Q$and $W^{}_K$.

Each entry $M_{i,j}$ in the calculated Map $M$ represents the attention weight of the $j^{\text{th}}$ text token for the $i^{\text{th}}$ spatial location. 
This mechanism allows the model to selectively amplify image features based on their relevance to the text, thereby reinforcing the alignment between the textual prompt and the visual output.
Furthermore, recent work~\cite{mao2023guided} has shown that, when a text prompt $c$ is fixed, the resulting cross-attention maps are entirely determined by the noise latent $z_t$. 
Since DDIM sampling is deterministic, the initial noise latent $z_T$ governs the entire denoising trajectory through the cross-attention layers, making it a crucial factor in controlling the generation outcome.

\vspace{5pt}
\subsection{Inverted Noise Latents in Domain Shift}

\begin{figure*}[t]
    \centering
    \includegraphics[width=0.95\linewidth]{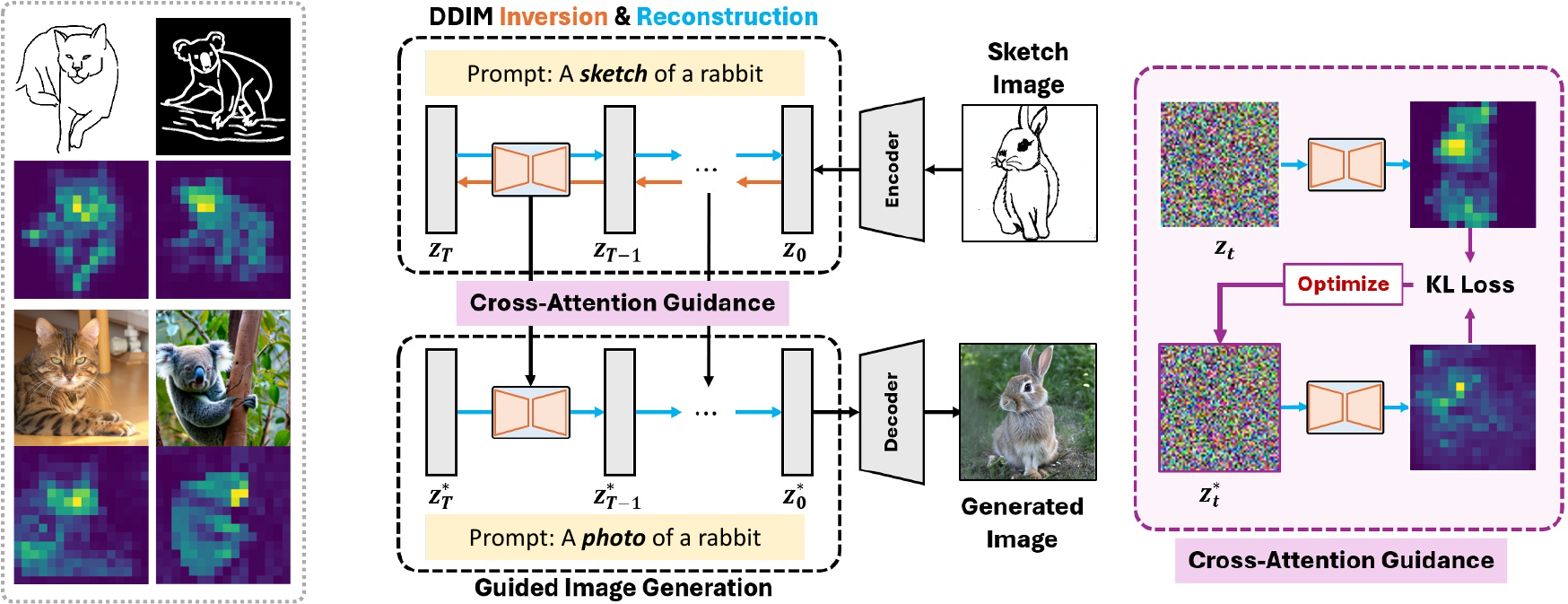}
    %\vspace{-15pt}
    \caption{
    Illustration of the proposed latent optimization pipeline and cross-attention visualization.
    As shown on the right side, we observe that the cross-attention maps remain robust to domain shifts when provided with prompts containing the domain information (see Sec.~\ref{subsec:bridging-gap}).
    For image generation, we obtain the inverted sketch noise latents $z_{T}$ through DDIM inversion.
    Next, we denoise the sketch latents using the source prompt $p_s$ to derive the attention maps corresponding to the sketch image.
    Finally, we employ randomly sampled initial noise latent $z_{T}^*$ alongside the target prompt $p_t$ to generate a new image. By utilizing KL loss, we facilitate the alignment of the cross-attention maps with those from the sketch.
    }
    \label{fig:pipeline}
    \vspace{-15pt}
\end{figure*}

\label{sec:noise_latent}
\subsubsection{Domain Gap among Inverted Noise Latents}
Despite their success in image editing, methods such as~\cite{tumanyan2023plug, hertz2022prompt} face limitations when applied to inputs outside the natural image domain—such as sketches. We observe that the inherent generation tendency of the initial noise latent obtained via DDIM inversion significantly restricts performance in such cases.

To investigate this issue, we conducted an experiment to examine whether the inverted noise distributions of natural and sketch images follow a standard Gaussian distribution. 
Specifically, we randomly selected 100 real images, along with 200 sketches (100 from the ImageNet-Sketch dataset and 100 from the Sketchy database), and applied 50 steps of DDIM inversion to each sample.
As shown in Fig.~\ref{fig:initial-generation-tendency}, inverted latents from natural images remain close to the standard Gaussian distribution.
In contrast, inverted sketch latents exhibit a much lower variance, despite having a similar mean, indicating a significant deviation from the natural image domain.
Furthermore, Fig.~\ref{fig:initial-generation-tendency} illustrates that when using sketch-inverted noise latents as the starting point for DDIM generation, the model fails to produce realistic images. This demonstrates that sketch-derived noise latents are less effective for generation compared to those from real images.
These findings highlight a key limitation: editing methods that rely on noise inverted from real images perform poorly on sketches due to this underlying distribution misalignment, thereby restricting their applicability in sketch-to-image tasks.

\subsubsection{Cross-Attention Robustness to Domain Shift}
\label{subsec:cross-attention-robustness}
To address the domain gap issue, we leverage the strong visual understanding capabilities of pre-trained diffusion models.
A key observation in our work is that pre-trained Stable Diffusion models exhibit notable domain invariance in its cross-attention maps.
Given an appropriate text prompt, the model can effectively localize regions in an image that correspond to the semantic concepts described in the prompt—regardless of whether the image is a sketch or a photo, as visualized in the right part of Fig.~\ref{fig:pipeline}. 
% add 

We present two sets of attention maps for visualization. 
The left column includes cross-attention maps for a real cat and a sketch of a cat, while the right column includes maps for a real koala and an edge map of a koala, shown in inverted black and white. 
In the reconstruction process of a koala sketch under the text prompt \verb|"a sketch of a koala"|, cross-attention layers try to identify regions in the image resembling the concept of a \verb|sketchy koala| and is eventually presented as the cross-attention map.
If we replace the sketch with a realistic image and substitute the word \verb|sketch|
in the prompt with the word \verb|photo|, the cross-attention layers will try to find regions in the image that resemble a \verb|real koala|.

Despite differences in appearance and even color inversion in the edge map, the attention maps behave consistently across sketches and photos, demonstrating their robustness to domain and style variations.
Motivated by this observation, we utilize cross-attention maps under different prompts to extract semantic information from the reference sketch, as detailed in Section~\ref{sec:extract}.
%This strategy allows us to bridge the domain gap without the need for additional training.

%\vspace{20pt}

%\vspace{-5pt}
\subsubsection{Bridging Domain Gap with Attention Maps}
\label{subsec:bridging-gap}

\begin{figure}[t]
    \centering
    \includegraphics[width=0.98\linewidth]{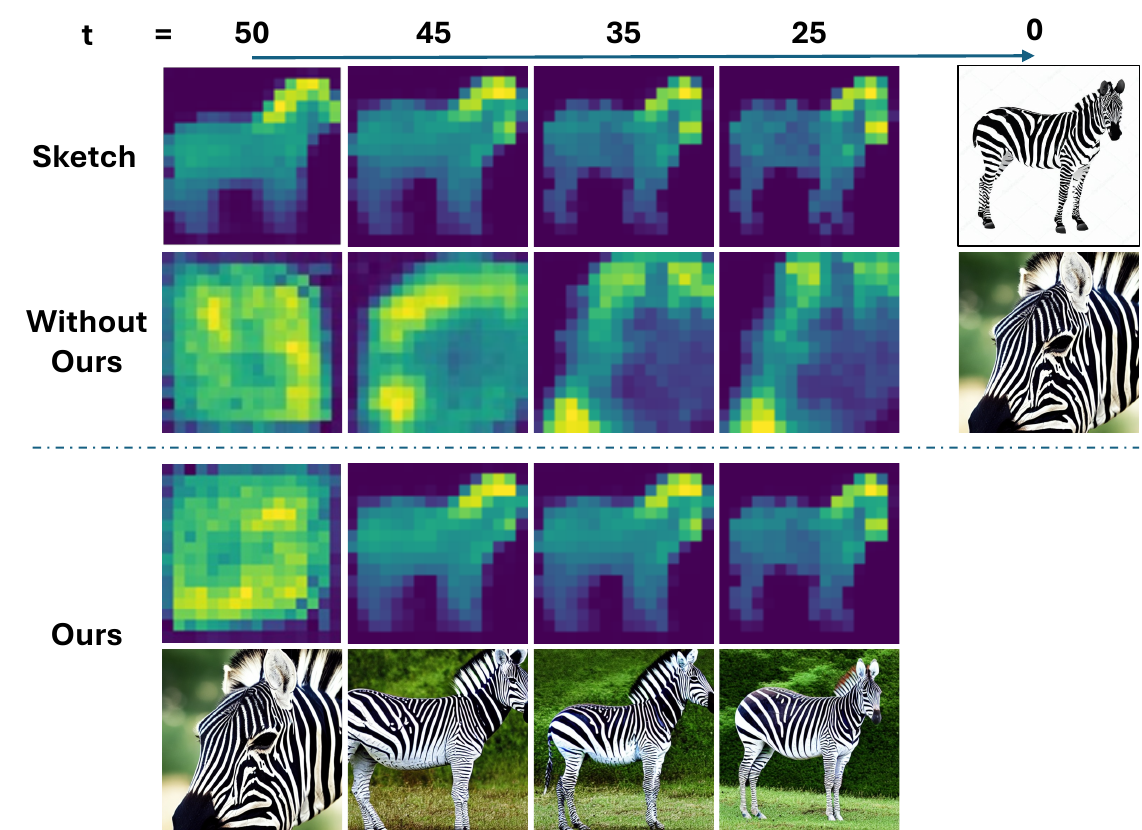}
    \vspace{-5pt}
    \caption{
    %Top: Visualization of cross-attention maps at time step $t$ for two processes: sketch reconstruction (first row) and image generation with a random seed $z$ (second row).
    %Bottom: Using the same random seed $z$, the third row shows how the cross-attention maps evolve when applying our proposed optimization technique up to the stopping time $t$. The corresponding generated images are presented in the fourth row.
    Top: Cross-attention maps at time step $t$ from sketch reconstruction (first row) and image generation with random seed $z$ (second row).
    Bottom: With the same seed $z$, the third row shows attention maps with our optimization, and the fourth row shows the corresponding generated images.}
    \label{fig:optimization-time-visualization}
    \vspace{-20pt}
\end{figure}

\begin{figure*}
    \centering
    \vspace{-10pt}
    \includegraphics[width=1.0\linewidth]{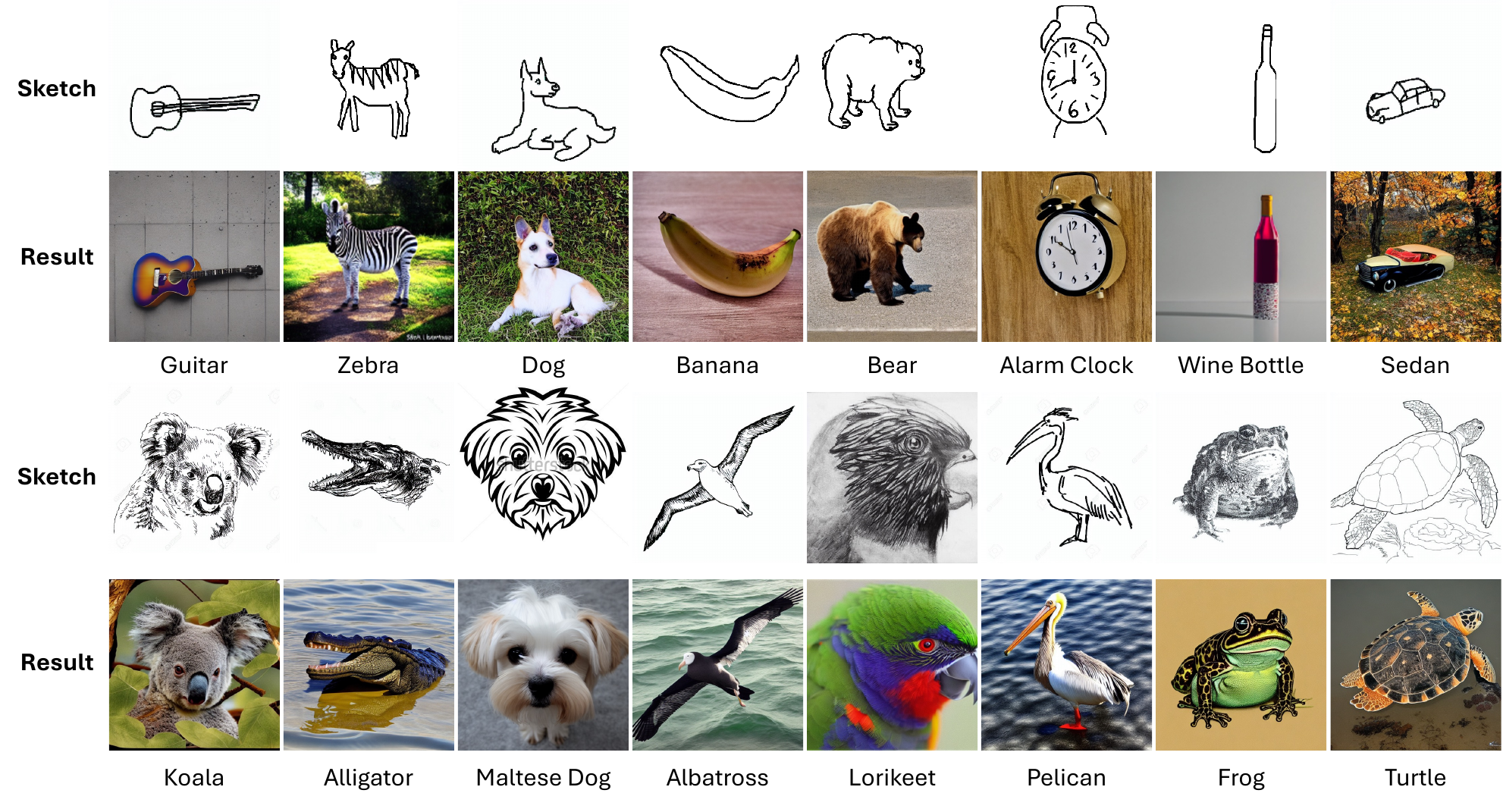} 
    \caption{
    Sketch to image translations on the Sketchy database~\cite{sketchy2016} (first row) and the ImageNet-Sketch dataset~\cite{wang2019learning} (second row). Our approach effectively translates these sketch images into realistic images. Even when the sketch is very scribbled, our method can still capture object features in the sketch guide and reproduce them in the generated image.} 
    \label{fig:results}
    \vspace{-15pt}
\end{figure*}

A straightforward approach to sketch-to-image translation might be directly substituting the cross-attention maps from a reference sketch image during the generation process. 
However, as discussed in Section~\ref{sec:cross-attention} and supported by~\cite{mao2023guided},  this strategy often causes inconsistencies with the layout information encoded in the spatial features derived from the initial noise latent, leading to suboptimal image quality.
To overcome this issue, we propose a more effective strategy: rather than replacing attention maps, we treat the cross-attention maps of the reference image as an optimization target, as detailed in Section~\ref{sec:optimize}.
%we ensure better alignment between the structural guidance from the sketch and the final generated image
 
%However, simply substituting the cross-attention maps often creates inconsistencies with the layout information encoded in the original spatial feature layers, leading to suboptimal image generation.
%To achieve the sketch-guided image generation, we propose to use the cross-attention maps of the reference image as the optimization objective to refine the noise latent directly, as we described in Sec.~\ref{sec:optimize}, instead of making a substitution of the attention maps.

\subsection{Guided Image Generation}
\label{sec:pipeline}

Based on previous discussions, we propose a novel pipeline for guided image generation. An overview of the entire process is illustrated in Fig.~\ref{fig:pipeline}. 
Additionally, Fig.~\ref{fig:optimization-time-visualization} presents visualizations of the extracted features at different time steps, along with the corresponding generated images when optimization is stopped at various stages.

\subsubsection{Feature Extraction from Reference Image}
\label{sec:extract}
To extract cross-attention maps from the reference sketch image, we first obtain a series of intermediate latents $\mathcal{Z}=\{ z_{0}, \dots, z_{T} \}$ by performing DDIM inversion on the given sketch $z_0$ using a pre-trained T2I Stable Diffusion. 
For each $z_i$ in $\mathcal{Z}$, we compute its cross-attention maps via the U-Net's cross-attention layers as $\mathcal{M}_i=\text{Cross Attention}(p,z_i)$, where $p$ denotes the embedding of the prompt \verb|"a sketch of a CLS"|. 
We preserve all cross-attention maps associated with the word \verb|"CLS"| into a collection $\mathcal{Q}=\{\mathcal{M}_i^\text{CLS}\}_{i=0,1,\dots, T}$. 

As discussed in Section~\ref{subsec:cross-attention-robustness}, cross-attention maps capture the spatial distribution of semantic features within the noisy latent.
By aligning the attention maps produced during image generation with those extracted from the reference sketch, we ensure that the semantic layout of the generated image closely matches that of the sketch.
Therefore, during the generation process, we use the extracted cross-attention maps $\mathcal{M}$ as optimization targets to refine our latent representation at each step.

\begin{figure*}[t]
    \centering
    \includegraphics[width=1\linewidth]{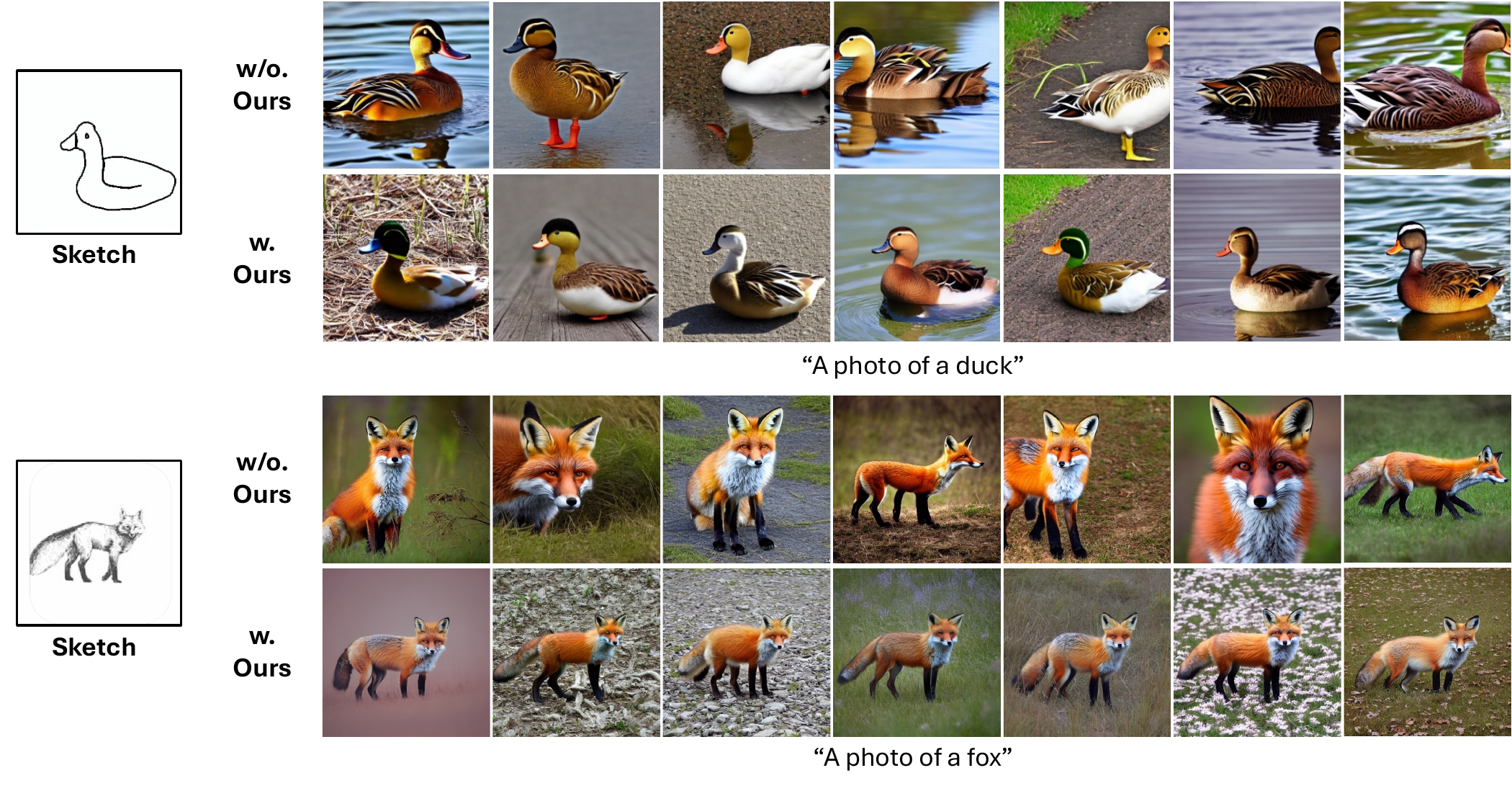}
    \vspace{-5pt}
    \caption{Sketch-to-image translations using different random seeds. Images within each column share the same random seed. While Stable Diffusion generates images with high randomness in layout, location, etc., our methods maintain visual variety from Stable Diffusion while adhering to the sketch structure.}
    \label{fig:random-seed-results}
    \vspace{-5pt}
\end{figure*}

\subsubsection{Image Generation with Latent Optimization}
\label{sec:optimize}

In this step, our goal is to generate a new image that adheres to the structural guidance of the sketch, building upon the insights presented in Sec.~\ref{sec:noise_latent}. 
To generate a photo-realistic image, we use the prompt \verb|"a photo of a CLS"| to guide the generation, where \verb|"CLS"| is the same category used in the feature extraction stage. 
The image generation process begins with a random noise $z_T^*$ sampled from a standard Gaussian distribution. 
To ensure the cross-attention maps of the generated image align with the target maps from the sketch, we employ cross-attention maps as guidance, progressively adjusting the intermediate latent so that its cross-attention maps closely match the target maps derived from the sketch.
We denote the cross-attention map at the $l$-th layer as $\mathcal{M}[l] \in [0, 1]^{\mathcal{N} \times \mathcal{N}}$. The similarity between the target and generated attention maps at the $t$-th step can then be calculated as follows:

\begin{equation}
    \mathcal{L}(\mathcal{M}_t^{\text{CLS}},\mathcal{M}_t^{*\text{CLS}}) = \sum_{l \in L} \mathcal{D}(\mathcal{M}_t^{\text{CLS}}[l], \mathcal{M}_t^{*\text{CLS}}[l])
    \label{eq:latent_optimization_loss}
\end{equation}
%where $\mathcal{D}$ is the symmetric Kullback–Leibler divergence:

\begin{equation}
    \mathcal{D}(x,y) = \mathcal{D}_{KL}(x\|y) + \mathcal{D}_{KL}(y\|x).
    \label{eq:KL_divergence}
\end{equation}
%Once the loss is calculated, we can introduce its gradient with respect to the input $z^*_t$ to obtain an optimized $\tilde{z}_t^*$ as follows:
\begin{equation}
\tilde{z}_{t}^{*} = z_{t}^{*} -
        \beta \cdot 
        \frac
            {{\|z_{t}^{*} - z_{t-1}^{*}\|_2}}
            {\|{\nabla}_{z_{t}^{*}}\mathcal{L}\|_2} 
        {{\nabla}_{z_{t}^{*}}\mathcal{L}}
\end{equation}
where $\beta$ controls the strength of the guidance.

\section{Experiments}
\label{sec:experiments}
We evaluate our proposed pipeline through two tasks, sketch-guided image generation and sketch-guided real image editing, as illustrated in Fig.~\ref{fig:teaser}.
We introduce the dataset and experimental setup in Sec.~\ref{subsec:dataset}, and our experimental results are presented in Sec.~\ref{subsec:result}.

\begin{figure*}[t]
    \centering
    \includegraphics[width=0.95\linewidth]{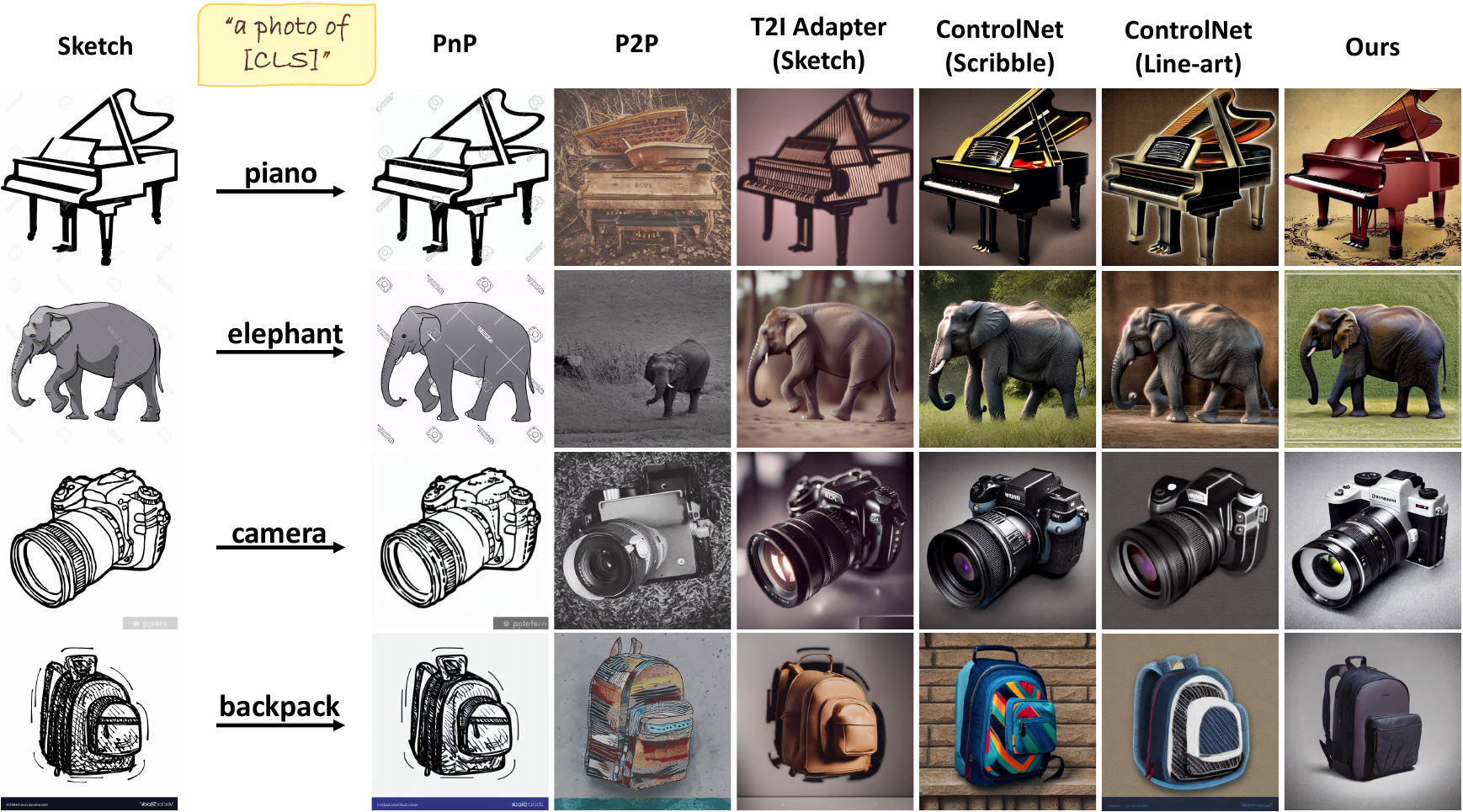}
    \caption{
    From left to right: the guidance sketch image, PnP~\cite{tumanyan2023plug}, P2P~\cite{hertz2022prompt}, T2I-Adapter~\cite{mou2023t2i} (with sketch adapter), ControlNet (scribble and line-art versions), and our results. PnP struggles to deviate from the input sketch, producing outputs that closely resemble the sketch but lack realism. P2P generates images that suffers from object misalignment. T2I-Adapter produces visually appealing results but lacks generalizability across diverse sketch types. ControlNet has better photorealism but still faces a trade-off between realistic appearance and strict adherence to the sketch structure.} 
    \label{fig:comparison}
\vspace{-10pt}
\end{figure*}

\subsection{Dataset and Experimental Setup}
\label{subsec:dataset}
%ImageNet-Sketch
We evaluate our methods on the Sketchy database~\cite{sketchy2016} and ImageNet-Sketch dataset~\cite{wang2019learning}.
The Sketchy database contains line-art sketches, their corresponding real images, and image class labels.
The sketch canvas has a resolution of $256\times256$, where each sketch undergoes the same scaling as its paired image within the database.
The ImageNet-Sketch dataset contains sketches and shares the same image class labels as ImageNet.
All images are resized to $512\times512$ for DDIM inversion to obtain initial noise images. 
Throughout the denoising process, we employ the prompt \verb|"A sketch of a CLS"|, wherein \verb|"CLS"| represents the class label of the associated sketch image.

%\subsection{Implementation Details}
%\label{subsec:implementation}
We use stable diffusion v1.5 with its default configuration as our baseline model.
We employ DDIM sampling with 50 steps for each image and set the classifier-free guidance scale to 7.5. %($T = 50$) 
Since the final steps of the denoising process have minimal influence on the overall layout, we apply our guidance only during the early stages.
As illustrated in Fig.~\ref{fig:optimization-time-visualization}, a few early steps help reposition the object toward the target area, while applying guidance for approximately half the denoising process refines both the contour and semantic content to closely align with the reference sketch. 
Empirically, applying guidance for the first 25 steps achieves the best performance.

\subsection{Result}
\label{subsec:result}

\subsubsection{Sketch-guided Image Generation}
\label{subsubsec:s2i}
%\textbf{Sketch-guided Image Generation}

In Fig.~\ref{fig:results}, we illustrate image results that demonstrate the capability of our pipeline to generate images across diverse sketch types.
In Fig.~\ref{fig:random-seed-results}, we further demonstrate that our method can produce diverse image results by varying the initial noise seed.
While the generation exhibit unpredictability regarding image layout, object structure, and location with Stable Diffusion, our pipeline maintains consistency in object placement and structure, even across different seeds, while still producing diverse outputs.

In our comparisons with other methods (Fig.~\ref{fig:comparison}), we focus on state-of-the-art baselines capable of utilizing sketches as inputs for image translation tasks. 
Specifically, we compare with (i) Plug-and-Play (PnP)~\cite{tumanyan2023plug}, (ii) Prompt-to-Prompt (P2P)~\cite{hertz2022prompt}, (ii) T2I-Adapter~\cite{mou2023t2i}, and (iv) ControlNet~\cite{zhang2023adding}. 
PnP and P2P are training-free methods commonly employed in text-guided image-to-image translation tasks, while T2I-Adapter and ControlNet are training-based.

In the sketch-to-image generation task, PnP struggles to deviate sufficiently from the guidance image, often producing results with noticeable visual artifacts. 
This limitation stems from its reliance on spatial feature injection and self-attention substitution derived from the source image.
Because the injected spatial features preserve semantic information from the guidance image, they hinder the transformation into a truly photo-realistic image.
P2P also demonstrates limited generative capability in sketch-to-image translation, as discrepancies between the object layout in the input sketch and the random initial noise hinder its performance.
T2I-Adapter demonstrates effectiveness only when the reference image closely matches the distribution of its training data.
Its performance becomes unstable when dealing with reference images in unseen styles.
Compared to the lightweight T2I-Adapter, ControlNet achieves higher realism. 
However, it still faces a trade-off between maintaining photorealism and accurately preserving the object structure defined in the sketch.
In contrast, our training-free approach effectively extracts meaningful features from the reference image and faithfully reconstructs them in a realistic style within the generated image. 
Despite not requiring training, our results are comparable to those of the training-based methods, demonstrating its potential as a efficient and flexible alternative.

\begin{table}[htbp]
    \centering
    \caption{Metrics: FID, IoU, and LPIPS}
    \vspace{-5pt}
    \begin{tabular}{lccc}
        \toprule
        Method & FID $\downarrow$ & IoU $\uparrow$ & LPIPS \\
        \midrule
        PnP                    & 69.2503 & 0.8289 & 0.3333 \\
        P2P                    & 46.6889 & 0.7081 & 0.4032 \\
        T2I Adapter (Sketch)    & 39.9186 & 0.7549 & 0.8036 \\
        %Controlnet (Lineart)     & 30.2533 & 0.8240 & 0.7413 \\
        ControlNet (Sketch)      & 27.4155 & 0.6488 & 0.8633 \\
        %Replacing Cross-attention   & 22.1670 & 0.4966 & 0.8682 \\
        Ours                   & 21.5380 & 0.6545 & 0.8165 \\
        \bottomrule
    \end{tabular}
    \vspace{-15pt}
    \label{tab:fid_iou_lpips}
\end{table}

\vspace{-5pt}
\subsubsection{Real Image Editing}

%\vspace{-5pt}
\begin{figure*}[t]
    \centering
    \includegraphics[width=0.98\linewidth]{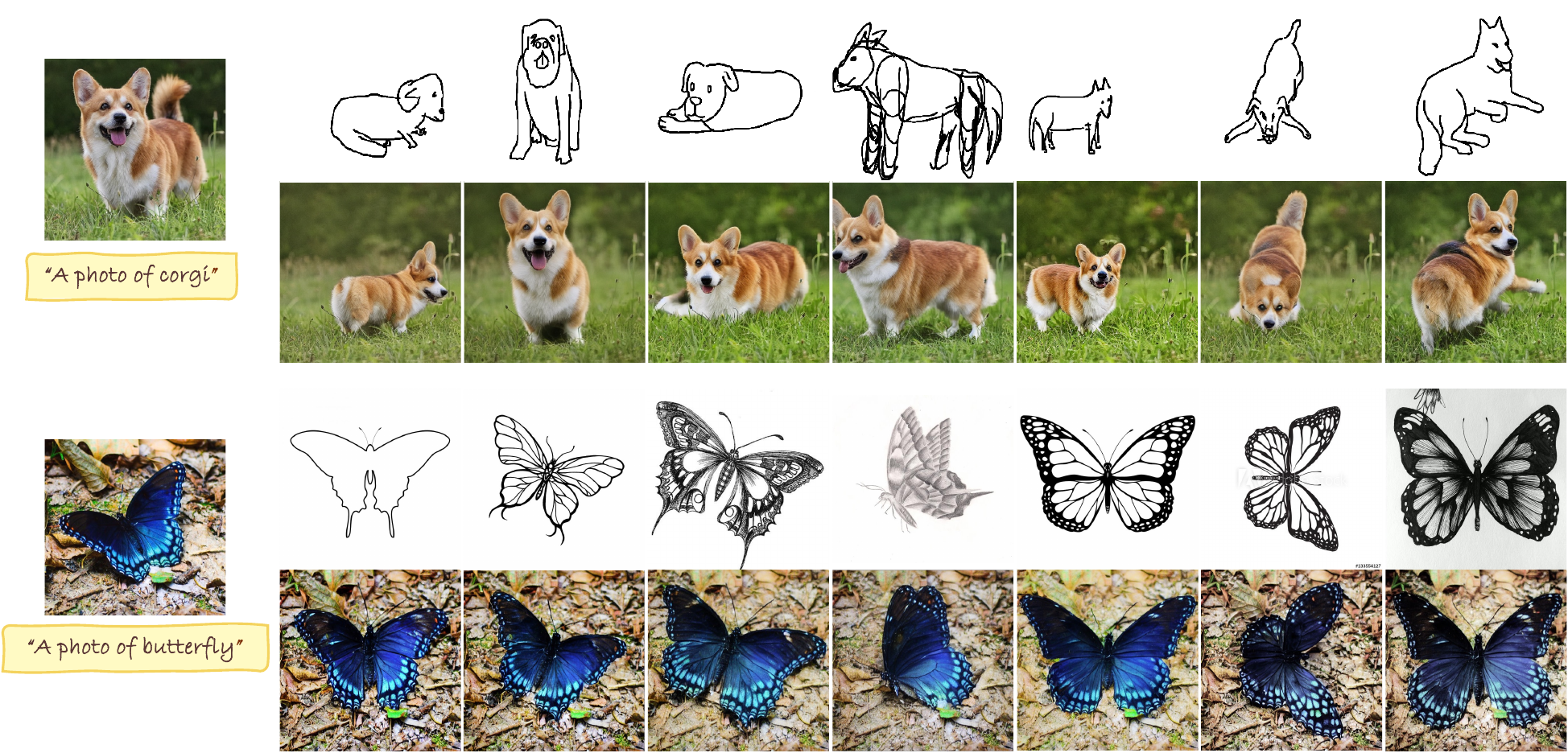}
    %\vspace{-5pt}
    \caption{
    Synthesis results using an exemplar image, a sketch reference, and a text prompt.  The illustrations demonstrate that combined with~\cite{cao2023masactrl}, we can achieve consistent image editing by blending the sketch image layout with the exemplar image's visual contents.}
    \label{fig:image-editing}
    %\vspace{-5pt}
\end{figure*}

To further validate performance, we conduct a quantitative evaluation using FID, IoU, and LPIPS on 10k sketches from the ImageNet-Sketch dataset, with ImageNet-1k as the reference for FID calculation.
For IoU, we use groundingDINO \cite{liu2024grounding} for object detection.
As shown in Table~1, our method achieves the lowest FID (21.54), indicating a high degree of photorealism and better alignment with the natural image distribution.

While our IoU (0.6545) and LPIPS (0.8165) are competitive, we note that these metrics do not fully reflect the perceptual quality in sketch-based generation.
For example, although PnP achieves the highest IoU (0.8289) and lowest LPIPS (0.3333), its FID (69.25) is substantially worse—implying over-reliance on sketch structure at the expense of realism.

Given a reference sketch, our method can effectively generate new images that preserve the layout and structure of the input sketch.
As shown in Fig.~\ref{fig:random-seed-results}, the visual characteristics and fine details of the generated images vary with different initial noise seeds.
A related study~\cite{cao2023masactrl} proposes injecting visual features from an exemplar image via spatial feature substitution in the self-attention blocks. 
Inspired by this, we integrate their method into our pipeline by performing feature substitution at each denoising step while simultaneously optimizing the latent based on cross-attention maps.
Specifically, we extract spatial features from the self-attention blocks of the exemplar image and substitute them into the generation process. 
This integration allows our method to preserve the structural layout from the sketch, while transferring the visual appearance from the exemplar image.
As a result, our pipeline can be extended to support real-image editing tasks guided by sketches.
As shown in Fig.~\ref{fig:image-editing}, the results demonstrate that our method can be effectively combined with~\cite{cao2023masactrl} to achieve independent control over both structural layout and visual style.

%\vspace{-5pt}
\section{Limitations and Discussion}
\label{sec:limitations}
%\vspace{-5pt}

Our method inherits certain limitations from Stable Diffusion, particularly in generating images that precisely match the desired structure. 
Although our method is capable of synthesizing images with layouts similar to the input sketch, minor deviations may still occur. 
This is because our approach relies heavily on the layout information encoded in cross-attention maps derived from the sketch. 
If the model fails to accurately interpret the sketch's structure or spatial arrangement, the resulting image may deviate from the intended composition. 

This issue is further compounded by the fact that Stable Diffusion is primarily trained on natural image domains, which limits its ability to interpret abstract or stylized sketches effectively. 
The limitation becomes particularly evident when processing highly abstract sketches, such as those with incomplete or unenclosed edges. 
While humans can intuitively fill in gaps and recognize the intended object, the model lacks such inference capabilities and may fail to reconstruct the correct layout.

%\begin{figure}[t]
 %   \centering
  %  \includegraphics[width=0.95\linewidth]{images/pdf/limitation.png}
   % \caption{Failure cases of the proposed method. Our pipeline struggles with abstract sketches with non-closed edges. It may also overlook some local geometry details or fail to understand the flip-over transformation of the input sketches.
    %It may also fail to generate images with correct geometry or accurately process flip-over transformed input sketches.
    %}
    %\label{fig:failure-cases}
%\end{figure}

%\vspace{10pt}

\section{Conclusion}

We present a novel training-free pipeline for the sketch-to-image generation task, which requires no model training or fine-tuning. 
By leveraging the cross-attention mechanism in pre-trained text-to-image diffusion models, our method effectively extracts layout and object structure from a sketch and utilizes these core features for guided image generation through latent optimization.

Our approach achieves a remarkable balance between preserving the spatial structure of the sketch and aligning with the semantic content of the text prompt. Furthermore, it can be integrated with other training-free methods, enabling both image variation generation and real-image editing without model fine-tuning.
Our work showcases the untapped potential of pre-trained text-to-image models, and we hope it will inspire future research in this direction.

{
    \small
    \bibliographystyle{ieeenat_fullname}
    \bibliography{main}
}


\end{document}